\documentclass[11pt]{article}
\usepackage{eacl2017}
\usepackage{times}
\usepackage{url}
\usepackage{latexsym,graphicx,algorithmic,amsmath}
\usepackage{multirow}
\usepackage{booktabs}
\usepackage{enumitem}
\usepackage[font={small}]{caption}
\usepackage[colorinlistoftodos]{todonotes}

\eaclfinalcopy
\def\eaclpaperid{156}

\title{A Strong Baseline for Learning Cross-Lingual Word Embeddings\\from Sentence Alignments}

\author{
 	    Omer Levy\thanks{These authors contributed equally to this work.}\\
 	    University of Washington\\
 	    Seattle, WA\\
 	    {\tt omerlevy@gmail.com}
 	  \And
 		Anders S{\o}gaard$^*$\\
   		University of Copenhagen\\
	   	Copenhagen, Denmark\\
		{\tt soegaard@di.ku.dk}
      \And
        Yoav Goldberg\\
        Bar-Ilan University\\
 	    Ramat-Gan, Israel\\
 	    {\tt yoav.goldberg@gmail.com}
}

\date{}

\begin{document}

\maketitle

\begin{abstract}
While cross-lingual word embeddings have been studied extensively in recent years, the qualitative differences between the different algorithms remain vague. We observe that whether or not an algorithm uses a particular feature set (sentence IDs) accounts for a significant performance gap among these algorithms. This feature set is also used by traditional alignment algorithms, such as IBM Model-1, which demonstrate similar performance to state-of-the-art embedding algorithms on a variety of benchmarks. Overall, we observe that different algorithmic approaches for utilizing the sentence ID feature space result in similar performance. This paper draws both empirical and theoretical parallels between the embedding and alignment literature, and suggests that adding additional sources of information, which go beyond the traditional signal of bilingual sentence-aligned corpora, may substantially improve cross-lingual word embeddings, and that future baselines should at least take such features into account.
\end{abstract}

\section{Introduction}

Cross-lingual word embedding algorithms try to represent the vocabularies of two or more languages in one common continuous vector space. These vectors can be used to improve monolingual word similarity \cite{Faruqui2014} or support cross-lingual transfer \cite{Gouws2015NAACL}. In this work, we focus on the second (cross-lingual) aspect of these embeddings, and try to determine what makes some embedding approaches better than others {\em on a set of translation-oriented benchmarks}. While cross-lingual word embeddings have been used for a variety of cross-lingual transfer tasks, we prefer evaluating on translation-oriented benchmarks, rather than across specific NLP tasks, since the translation setting allows for a cleaner examination of cross-lingual similarity. Another important delineation of this work is that we focus on algorithms that rely on {\em sentence-aligned data}; in part, because these algorithms are particularly interesting for low-resource languages, but also to make our analysis and comparison with alignment algorithms more focused.

We observe that the top performing embedding algorithms share the same underlying feature space -- sentence IDs -- while their different algorithmic approaches seem to have a negligible impact on performance. We also notice that several statistical alignment algorithms, such as IBM Model-1 \cite{Brown1993}, operate under the same data assumptions. Specifically, we find that using the translation probabilities learnt by Model-1 as the cross-lingual similarity function (in place of the commonly-used cosine similarity between word embeddings) performs on-par with state-of-the-art cross-lingual embeddings on word alignment and bilingual dictionary induction tasks. In other words, as long as the similarity function is based on the sentence ID feature space and the embedding/alignment algorithm itself is not too na\"{i}ve, the actual difference in performance between different approaches is marginal.

This leads us to revisit another statistical alignment algorithm from the literature that uses the same sentence-based signal -- the Dice aligner \cite{Och2003}. We first observe that the vanilla Dice aligner is significantly outperformed by the Model-1 aligner. We then recast Dice as the dot-product between two word vectors (based on the sentence ID feature space), which allows us to generalize it, resulting in an embedding model that is as effective as Model-1 and other sophisticated state-of-the-art embedding methods, but takes a fraction of the time to train.

Existing approaches for creating cross-lingual word embeddings are typically restricted to training bilingual embeddings, mapping exactly two languages onto a common space. We show that our generalization of the Dice coefficient can be augmented to jointly train \emph{multi}-lingual embeddings for any number of languages. We do this by leveraging the fact that the space of sentence IDs is shared among all languages in the parallel corpus; the verses of the Bible, for example, are identical across all translations. Introducing this multi-lingual signal shows a significant performance boost, which eclipses the variance in performance among pre-existing embedding algorithms.

\paragraph{Contributions} We first establish the importance of the sentence ID feature space for cross-lingual word embedding algorithms through experiments across several translation-oriented benchmarks. We then compare cross-lingual word embedding algorithms to traditional word alignment algorithms that also rely on sentence ID signals. We show that a generalization of one of these, the Dice aligner, is a very strong baseline for cross-lingual word embedding algorithms, performing better than several state-of-the-art algorithms, especially when exploiting a multi-lingual signal. Our code and data are publicly available.\footnote{\url{bitbucket.org/omerlevy/xling_embeddings}}

\section{Background: Cross-lingual Embeddings}
\label{sec:background}

Previous approaches to cross-lingual word embeddings can be divided into three categories, according to assumptions on the training data. The first category assumes \emph{word-level alignments}, in the form of bilingual dictionaries \cite{Mikolov2013Crosslingual,Xiao2014} or automatically produced word alignments \cite{Klementiev2012,Zou2013,Faruqui2014}. Sizable bilingual dictionaries are not available for many language pairs, and the quality of automatic word alignment greatly affects the quality of the embeddings. It is also unclear whether the embedding process provides significant added value beyond the initial word alignments \cite{Zou2013}. We therefore exclude these algorithms for this study, also in order to focus our analysis and make the comparison with traditional word alignment algorithms more straightforward.

The second category makes a much weaker assumption, \emph{document-level alignments}, and uses comparable texts in different languages (not necessarily translations) such as Wikipedia articles or news reports of the same event. Algorithms in this category try to leverage massive amounts of data to make up for the lack of lower-level alignments \cite{Soegaard2015,Vulic2016}.

Algorithms in the third category take the middle ground; they use \emph{sentence-level alignments}, common in legal translations and religious texts. Also known as ``parallel corpora'', sentence-aligned data maps each sentence (as a whole) to its translation. We focus on this third category, because it does not require the strict assumption of word-aligned data (which is difficult to obtain), while still providing a cleaner and more accurate signal than document-level alignments (which have been shown, in monolingual data, to capture mainly syntagmatic relations \cite{Sahlgren2006}). In \S6, we provide evidence to the hypothesis that sentence-aligned data is indeed far more informative than document-aligned data.

Algorithms that rely on sentence-aligned data typically create intermediate sentence representations from each sentence's constituent words. \newcite{Hermann2014} proposed a deep neural model, BiCVM, which compared the two sentence representations at the final layer, while \newcite{Chandar2014} proposed a shallower autoencoder-based model, representing both source and target language sentences as the same intermediate sentence vector. Recently, a simpler model, BilBOWA \cite{Gouws2015Bilbowa}, showed similar performance \emph{without} using a hidden sentence-representation layer, giving it a dramatic speed advantage over its predecessors. BilBOWA is essentially an extension of skip-grams with negative sampling (SGNS) \cite{Mikolov2013Distributed}, which simultaneously optimizes each word's similarity to its inter-lingual context (words that appeared in the aligned target language sentence) and its intra-lingual context (as in the original monolingual model). \newcite{Luong2015NAACL} proposed a similar SGNS-based model over the same features.

We study which factors determine the success of cross-lingual word embedding algorithms that use sentence-aligned data, and evaluate them against baselines from the statistical machine translation literature that incorporate the same data assumptions. We go on to generalize one of these, the Dice aligner, showing that one variant is a much stronger baseline for cross-lingual word embedding algorithms than standard baselines.

Finally, we would like to point out the work of \newcite{Upadhyay2016}, who studied how different data assumptions affect embedding quality in both monolingual and cross-lingual tasks. Our work focuses on one specific data assumption (sentence-level alignments) and only on cross-lingual usage. This more restricted setting allows us to: (a) compare embeddings to alignment algorithms, (b) decouple the feature space from the algorithm, and make a more specific observation about the contribution of each component to the end result. In that sense, our findings complement those of \newcite{Upadhyay2016}.

\section{Which Features Make Better Cross-lingual Embeddings?}

We group state-of-the-art cross-lingual embedding algorithms according to their feature sets, and compare their performance on two cross-lingual benchmarks: word alignment and bilingual dictionary induction. In doing so, we hope to learn which features are more informative.

\subsection{Features of Sentence-aligned Data}
\label{sec:features}

We observe that cross-lingual embeddings typically use parallel corpora in one of two ways:

\paragraph{Source + Target Language Words}
Each word $w$ is represented using all the other words that appeared with it in the same sentence (source language words) \emph{and} all the words that appeared in target language sentences that were aligned to sentences in which the word $w$ appeared (target language words). This representation also stores the number of times each pair of word $w$ and feature (context) word $f$ co-occurred.

These features are analogous to the ones used by \newcite{Vulic2016} for document-aligned data, and can be built in a similar manner: create a pseudo-bilingual sentence from each aligned sentence, and for each word in question, consider all the other words in this sentence as its features. BilBOWA \cite{Gouws2015Bilbowa} also uses a similar set of features, but restricts the source language words to those that appeared within a certain distance from the word in question, and defines a slightly different interaction with target language words.

\paragraph{Sentence IDs}
Here, each word is represented by the set of sentences in which it appeared, indifferent to the number of times it appeared in each one. This feature set is also indifferent to the word ordering within each sentence. This approach is implicitly used by \newcite{Chandar2014}, who encode the bag-of-words representations of parallel sentences into the same vector. Thus, each word is not matched directly to another word, but rather used to create the sentence's language-independent representation. \newcite{Soegaard2015} use similar features, document IDs, for leveraging comparable Wikipedia articles in different languages. In \S\ref{sec:data-paradigms} we show that when using sentence IDs, even a small amount of sentence-aligned data is more powerful than a huge amount of comparable documents.

\subsection{Experiment Setup}

\paragraph{Algorithms}
We use the four algorithms mentioned in \S\ref{sec:features}: BilBOWA \cite{Gouws2015Bilbowa}, BWE-SkipGram \cite{Vulic2016}, Bilingual Autoencoders \cite{Chandar2014}, and Inverted Index \cite{Soegaard2015}. While both BWE-SkipGram and Inverted Index were originally trained on document-aligned data, in this work, we apply them to sentence-aligned data. 

\paragraph{Data}
\newcite{Christodouloupoulos2015} collected translations of the Bible (or parts of it) in over 100 languages, naturally aligned by book, chapter, and verse (31,102 verses in total).\footnote{\url{homepages.inf.ed.ac.uk/s0787820/bible/}} This corpus allows us to evaluate methods across many different languages, while controlling for the training set's size. The corpus was decapitalized and tokenized using white spaces after splitting at punctuation.

\paragraph{Benchmarks}
We measure the quality of each embedding using both manually annotated word alignment datasets and bilingual dictionaries. We use 16 manually annotated word alignment datasets -- Hansards\footnote{\url{www.isi.edu/natural-language/download/hansard/}} and data from four other sources  \cite{Graca2008,Lambert:ea:05,Mihalcea:Pedersen:03,Holmqvist2011,Cakmak:ea:12} -- as well as 16 bilingual dictionaries from Wiktionary.

In the word alignment benchmark, each word in a given source language sentence is aligned with the most similar target language word from the target language sentence -- this is exactly the same greedy decoding algorithm that is implemented in IBM Model-1 \cite{Brown1993}. If a source language word is out of vocabulary, it is not aligned with anything, whereas target language out-of-vocabulary words are given a default minimal similarity score, and never aligned to any candidate source language word in practice. We use the inverse of alignment error rate (1-AER) as described in \newcite{Koehn2010} to measure performance, where higher scores mean better alignments.

High quality, freely available, manually annotated word alignment datasets are rare, especially for non-European languages. We therefore include experiments on bilingual dictionary induction. We obtain bilingual dictionaries from Wiktionary for five non-Indo-European languages, namely: Arabic, Finnish, Hebrew, Hungarian, and Turkish (all represented in the Edinburgh Bible Corpus). We emphasize that unlike most previous work, we experiment with finding translation equivalents of all words and do not filter the source and target language words by part of speech. We use precision-at-one (P@1), essentially selecting the closest target-language word to the given source-language word as the translation of choice. This often means that 100\% precision is unattainable, since many words have multiple translations.

\paragraph{Hyperparameters}
\newcite{Levy2015} exposed a collection of hyperparameters that affect the performance of monolingual embeddings. We assume that the same is true for cross-lingual embeddings, and use their recommended settings across all algorithms (where applicable). Specifically, we used 500 dimensions for every algorithm, context distribution smoothing with $\alpha = 0.75$ (applicable to BilBOWA and BWE-SkipGram), the symmetric version of SVD (applicable to Inverted Index), and run iterative algorithms for 100 epochs to ensure convergence (applicable to all algorithms except Inverted Index). For BilBOWA's monolingual context window, we used the default of 5. Similarity is always measured by the vectors' cosine. Most importantly, we use a shared vocabulary, consisting of every word that appeared at least twice in the corpus (tagged with language ID). While hyperparameter tuning could admittedly affect results, we rarely have data for reliably tuning hyperparameters for truly low-resource languages.

\begin{table*}[t]
\begin{center}
\scalebox{0.8}{
\begin{tabular}{| c | c | c c || c c | c c | }
\hline
& & & & \multicolumn{2}{|c|}{\multirow{2}{*}{\bf Source+Target Words}} & \multicolumn{2}{c|}{\multirow{2}{*}{\bf Sentence IDs}} \\
& & & & & & & \\
\cline{5-8}
& & & & \multirow{2}{*}{BilBOWA} & BWE & Bilingual & Inverted \\
& & & & & SkipGram & Autoencoders & Index \\
\hline
\hline
\multirow{16}{*}{\rotatebox[origin=c]{90}{\textbf{\large{Word Alignment (1-AER)}}}} & \multirow{6}{*}{\small {\sc Gra\c{c}a}} & en & fr & .3653 & .3538 & \textbf{.4376} & .3499 \\
& & fr & en & .3264 & .3676 & \textbf{.4488} & .3995 \\
& & en & es & .2723 & .3156 & \textbf{.5000} & .3443 \\
& & es & en & .2953 & .3740 & \textbf{.5076} & .4545 \\
& & en & pt & .3716 & .3983 & \textbf{.4449} & .3263 \\
& & pt & en & .3949 & .4272 & \textbf{.4474} & .3902 \\
\cline{2-8}
& \multirow{2}{*}{\small {\sc Hansards}} & en & fr & .3189 & .3109 & \textbf{.4083} & .3336 \\
& & fr & en & .3206 & .3314 & \textbf{.4218} & .3749 \\
\cline{2-8}
& \multirow{2}{*}{\small {\sc Lambert}} & en & es & .1576 & .1897 & \textbf{.2960} & .2268 \\
& & es & en & .1617 & .2073 & \textbf{.2905} & .2696 \\
\cline{2-8}
& \multirow{2}{*}{\small {\sc Mihalcea}} & en & ro & .1621 & .1848 & \textbf{.2366} & .1951 \\
& & ro & en & .1598 & .2042 & \textbf{.2545} & .2133 \\
\cline{2-8}
& \multirow{2}{*}{\small {\sc Holmqvist}} & en & sv & .2092 & .2373 & \textbf{.2746} & .2357 \\
& & sv & en & .2121 & .2853 & \textbf{.2994} & .2881 \\
\cline{2-8}
& \multirow{2}{*}{\small {\sc Cakmak}} & en & tr & .1302 & .1547 & \textbf{.2256} & .1731 \\
& & tr & en & .1479 & .1571 & .2661 & \textbf{.2665} \\
\hline
\multirow{16}{*}{\rotatebox[origin=c]{90}{\textbf{\large{Dictionary Induction (P@1)}}}} & \multirow{16}{*}{\small {\textsc{Wiktionary}}} & en & fr & .1096 & .2176 & .2475 & \textbf{.3125} \\
& & fr & en & .1305 & .2358 & .2762 & \textbf{.3466} \\
& & en & es & .0630 & .1246 & .2738 & \textbf{.3135} \\
& & es & en & .0650 & .1399 & .3012 & \textbf{.3574} \\
& & en & pt & .1384 & \textbf{.3869} & .3281 & .3866 \\
& & pt & en & .1573 & .4119 & .3661 & \textbf{.4190} \\
& & en & ar & .0385 & \textbf{.1364} & .0995 & \textbf{.1364} \\
& & ar & en & .0722 & .2408 & .1958 & \textbf{.2825} \\
& & en & fi & .0213 & .1280 & .0887 & \textbf{.1367} \\
& & fi & en & .0527 & .1877 & .1597 & \textbf{.2477} \\
& & en & he & .0418 & \textbf{.1403} & .0985 & .1284 \\
& & he & en & .0761 & .1791 & .1701 & \textbf{.2179} \\
& & en & hu & .0533 & \textbf{.2299} & .1679 & .2182 \\
& & hu & en & .0810 & .2759 & .2234 & \textbf{.3204} \\
& & en & tr & .0567 & .2207 & .1770 & \textbf{.2245} \\
& & tr & en & .0851 & .2598 & .2069 & \textbf{.2835} \\
\hline
\hline
\multicolumn{4}{|c||}{\textbf{Average}*} & .1640 & .2505 & .2856 & \textbf{.2867} \\
\multicolumn{4}{|c||}{\textbf{Top 1}} & 0 & 3.5 & \textbf{15} & 13.5 \\
\hline
\end{tabular}
}
\end{center}
\caption{The performance of four state-of-the-art cross-lingual embedding methods. * Averages across two different metrics.}
\label{tab:results_sota}
\vspace{-5pt}
\end{table*}

\subsection{Results}

Table~\ref{tab:results_sota} shows that the two algorithms based on the sentence-ID feature space perform consistently better than those using source+target words. We suspect that the source+target feature set might be capturing more information than is actually needed for translation, such as syntagmatic or topical similarity between words (e.g. ``dog'' $\sim$ ``kennel''). This might be distracting for cross-lingual tasks such as word alignment and bilingual dictionary induction. Sentence ID features, on the other hand, are simpler, and might therefore contain a cleaner translation-oriented signal.

It is important to state that, in absolute terms, these results are quite poor. The fact that the best inverse AER is around 50\% calls into question the ability to actually utilize these embeddings in a real-life scenario. While one may suggest that this is a result of the small training dataset (Edinburgh Bible Corpus), previous work (e.g. \cite{Chandar2014}) used an even smaller dataset (the first 10K sentences in Europarl \cite{Europarl}). To ensure that our results are not an artifact of the Edinburgh Bible Corpus, we repeated our experiments on the full Europarl corpus (180K sentences) for a subset of languages (English, French, and Spanish), and observed similar trends. As this is a comparative study focused on analyzing the qualitative differences between algorithms, we place the issue of low absolute performance aside for the moment, and reopen it in \S\ref{sec:polyglot}.

\begin{table*}[t]
\begin{center}
\scalebox{0.8}{
\begin{tabular}{| c | c | c c || c c | c c | }
\hline
& & & & \multicolumn{2}{|c|}{\multirow{2}{*}{\bf Embeddings}} & \multicolumn{2}{c|}{\bf Alignment} \\
& & & & & & \multicolumn{2}{c|}{\bf Algorithms} \\
\cline{5-8}
& & & & Bilingual & Inverted & \multirow{2}{*}{Dice} & IBM \\
& & & & Autoencoders & Index & & Model-1 \\
\hline
\hline
\multirow{16}{*}{\rotatebox[origin=c]{90}{\textbf{\large{Word Alignment (1-AER)}}}} & \multirow{6}{*}{\small {\sc Gra\c{c}a}} & en & fr & \textbf{.4376} & .3499 & .3355 & .4263 \\
& & fr & en & \textbf{.4488} & .3995 & .3470 & .4248 \\
& & en & es & \textbf{.5000} & .3443 & .3919 & .4251 \\
& & es & en & \textbf{.5076} & .4545 & .3120 & .4243 \\
& & en & pt & .4449 & .3263 & .3569 & \textbf{.4729} \\
& & pt & en & .4474 & .3902 & .3598 & \textbf{.4712} \\
\cline{2-8}
& \multirow{2}{*}{\small {\sc Hansards}} & en & fr & .4083 & .3336 & .3614 & \textbf{.4360} \\
& & fr & en & .4218 & .3749 & .3663 & \textbf{.4499} \\
\cline{2-8}
& \multirow{2}{*}{\small {\sc Lambert}} & en & es & \textbf{.2960} & .2268 & .2057 & .2400 \\
& & es & en & \textbf{.2905} & .2696 & .1947 & .2443 \\
\cline{2-8}
& \multirow{2}{*}{\small {\sc Mihalcea}} & en & ro & \textbf{.2366} & .1951 & .2030 & .2335 \\
& & ro & en & \textbf{.2545} & .2133 & .1720 & .2214 \\
\cline{2-8}
& \multirow{2}{*}{\small {\sc Holmqvist}} & en & sv & .2746 & .2357 & .2435 & \textbf{.3405} \\
& & sv & en & .2994 & .2881 & .2541 & \textbf{.3559} \\
\cline{2-8}
& \multirow{2}{*}{\small {\sc Cakmak}} & en & tr & .2256 & .1731 & .2285 & \textbf{.3154} \\
& & tr & en & .2661 & .2665 & .2458 & \textbf{.3494} \\
\hline
\multirow{16}{*}{\rotatebox[origin=c]{90}{\textbf{\large{Dictionary Induction (P@1)}}}} & \multirow{16}{*}{{\small \textsc{Wiktionary}}} & en & fr & .2475 & \textbf{.3125} & .1104 & .1791 \\
& & fr & en & .2762 & \textbf{.3466} & .1330 & .1816 \\
& & en & es & .2738 & \textbf{.3135} & .1072 & .0903 \\
& & es & en & .3012 & \textbf{.3574} & .1417 & .1131 \\
& & en & pt & .3281 & \textbf{.3866} & .1384 & .3779 \\
& & pt & en & .3661 & .4190 & .1719 & \textbf{.4358} \\
& & en & ar & .0995 & \textbf{.1364} & .0449 & .1316 \\
& & ar & en & .1958 & .2825 & .0610 & \textbf{.2873} \\
& & en & fi & .0887 & \textbf{.1367} & .0423 & .1340 \\
& & fi & en & .1597 & \textbf{.2477} & .0463 & .2394 \\
& & en & he & .0985 & \textbf{.1284} & .0358 & .1224 \\
& & he & en & .1701 & \textbf{.2179} & .0328 & .2000 \\
& & en & hu & .1679 & .2182 & .0569 & \textbf{.2219} \\
& & hu & en & .2234 & \textbf{.3204} & .0737 & .2985 \\
& & en & tr & .1770 & \textbf{.2245} & .0406 & .1985 \\
& & tr & en & .2069 & .2835 & .0820 & \textbf{.3073} \\
\hline
\hline
\multicolumn{4}{|c||}{\textbf{Average}} & 0.2856 & 0.2867 & 0.1843 & \textbf{0.2922} \\
\multicolumn{4}{|c||}{\textbf{Top 1}} & 8 & \textbf{12} & 0 & \textbf{12} \\
\hline
\end{tabular}
}
\end{center}
\caption{The performance of embedding and alignment methods based on the sentence ID feature set.}
\label{tab:results_alignment}
\end{table*}

\section{Comparing Cross-lingual Embeddings with Traditional Alignment Methods}

Sentence IDs are not unique to modern embedding methods, and have been used by statistical machine translation from the very beginning. In particular, the Dice coefficient \cite{Och2003}, which is often used as a baseline for more sophisticated alignment methods, measures the cross-lingual similarity of words according to the number of aligned sentences in which they appeared. IBM Model-1 \cite{Brown1993} also makes exactly the same data assumptions as other sentence-ID methods. It therefore makes sense to use Dice similarity and the translation probabilities derived from IBM Model-1 as baselines for cross-lingual embeddings that use sentence IDs.

From Table~\ref{tab:results_alignment} we learn that the existing embedding methods are not really better than IBM Model-1. In fact, their average performance is even slightly lower than Model-1's. Although Bilingual Autoencoders, Inverted Index, and Model-1 reflect entirely different algorithmic approaches (respectively: neural networks, matrix factorization, and EM), the overall difference in performance seems to be rather marginal. This suggests that the main performance factor is not the algorithm, but the feature space: sentence IDs.

However, Dice also relies on sentence IDs, yet its performance is significantly worse. We suggest that Dice uses the sentence-ID feature set na\"{i}vely, resulting in degenerate performance with respect to the other methods. In the following section, we analyze this shortcoming and show that generalizations of Dice actually do yield similar performance Model-1 and other sentence-ID methods.


\section{Generalized Dice}

In this section, we show that the Dice coefficient \cite{Och2003} can be seen as the dot-product between two word vectors represented over the sentence-ID feature set. After providing some background, we demonstrate the mathematical connection between Dice and word-feature matrices. We then introduce a new variant of Dice, SID-SGNS, which performs on-par with Model-1 and the other embedding algorithms. This variant is able to seamlessly leverage the multi-lingual nature of sentence IDs, giving it a small but significant edge over Model-1.

\subsection{Word-Feature Matrices}
\label{sec:matrices}

In the word similarity literature, it is common to represent words as real-valued vectors and compute their ``semantic'' similarity with vector similarity metrics, such as the cosine of two vectors. These word vectors are traditionally derived from sparse word-feature matrices, either by using the matrix's rows as-is (also known as ``explicit'' representation) or by inducing a lower-dimensional representation via matrix factorization \cite{Turney2010}. Many modern methods, such as those in word2vec \cite{Mikolov2013Distributed}, also create vectors by factorizing word-feature matrices, only without representing these matrices explicitly.

Formally, we are given a vocabulary of words $V_W$ and a feature space (``vocabulary of features'') $V_F$. These features can be, for instance, the set of sentences comprising the corpus. We then define a matrix $M$ of $|V_W|$ rows and $|V_F|$ columns. Each entry in $M$ represents some statistic pertaining to that combination of word and feature. For example, $M_{w,f}$ could be the number of times the word $w$ appeared in the document $f$.

The matrix $M$ is typically processed into a ``smarter'' matrix that reflects the strength of association between each given word $w$ and feature $f$. We present three common association metrics: $L_1$ row normalization (Equation~(\ref{eq:l1})), Inverse Document Frequency (IDF, Equation~(\ref{eq:idf})), and Pointwise Mutual Information (PMI, Equation~(\ref{eq:ppmi})). The following equations show how to compute their respective matrices:
\begin{equation}
\scalebox{1.0}{$
M^{L_1}_{w,f} = \frac{I(w,f)}{I(w,*)}
$}
\label{eq:l1}
\end{equation}
\begin{equation}
\scalebox{1.0}{$
M^{IDF}_{w,f} = \log\frac{|V_F|}{I(w,*)}
$}
\label{eq:idf}
\end{equation}
\begin{equation}
\scalebox{1.0}{$
M^{PMI}_{w,f} = \log\frac{\#(w,f) \cdot \#(*, *)}{\#(w, *) \cdot \#(*, f)}
$}
\label{eq:ppmi}
\end{equation}
where $\#(\cdot, \cdot)$ is the co-occurrence count function, $I(\cdot, \cdot)$ is the co-occurrence indicator function, and $*$ is a wildcard.\footnote{A function with a wildcard should be interpreted as the sum of all possible instantiations, e.g. $I(w, *) = \sum_x I(w, x)$.}

To obtain word vectors of lower dimensionality ($V_F$ may be huge), the processed matrix is then decomposed, typically with SVD. An alternative way to create low-dimensional word vectors without explicitly constructing $M$ is to use the negative sampling algorithm (SGNS) \cite{Mikolov2013Distributed}.\footnote{For consistency with prior art, we refer to this algorithm as SGNS (skip-grams with negative sampling), even when it is applied without the skip-gram feature model.} This algorithm factorizes $M^{PMI}$ using a weighted non-linear objective \cite{Levy2014NIPS}.

\subsection{Reinterpreting the Dice Coefficient}
\label{sec:dice}

\noindent In statistical machine translation, the Dice coefficient is commonly used as a baseline for word alignment \cite{Och2003}. Given sentence-aligned data, it provides a numerical measure of how likely two words -- a source-language word $w_s$ and a target-language word $w_t$ -- are each other's translation:
\begin{equation}
\scalebox{1.0}{$
Dice(w_s, w_t) = \frac{2 \cdot S(w_s, w_t)}{S(w_s, *) \cdot S(*, w_t)}
$}
\label{eq:dice}
\end{equation}
where $S(\cdot, \cdot)$ is the number of aligned sentences in the data where both arguments occurred.

We claim that this metric is mathematically equivalent to the dot-product of two $L_1$-normalized sentence-ID word-vectors, multiplied by 2. In other words, if we use the combination of sentence-ID features and $L_1$-normalization to create our word vectors, then for any $w_s$ and $w_t$:
\begin{equation}
\scalebox{1.0}{$
w_s \cdot w_t = \frac{Dice(w_s, w_t)}{2}
$}
\end{equation}

To demonstrate this claim, let us look at the dot-product of $w_s$ and $w_t$:
\begin{equation}
\scalebox{1.0}{$
w_s \cdot w_t = \sum_i \left( \frac{I(w_s,i)}{I(w_s,*)} \cdot \frac{I(w_t,i)}{I(w_t,*)} \right)
$}
\label{eq:proof1}
\end{equation}
where $i$ is the index of an aligned sentence. Since $I(w_s,*) = S(w_s,*)$ and $I(w_t, *) = S(*,w_t)$, and both are independent of $i$, we can rewrite the equation as follows:
\begin{equation}
\scalebox{1.0}{$
w_s \cdot w_t = \frac{\sum_i I(w_s,i) \cdot I(w_t,i)}{S(w_s,*) \cdot S(*,w_t)}
$}
\label{eq:proof2}
\end{equation}
Since $I(w, i)$ is an indicator function of whether the word $w$ appeared in sentence $i$, it stands to reason that the product $I(w_s, i) \cdot I(w_t, i)$ is an indicator of whether both $w_s$ and $w_t$ appeared in $i$. Ergo, the numerator of Equation~(\ref{eq:proof2}) is exactly the number of aligned sentences in which both $w_s$ and $w_t$ occurred: $S(w_s, w_t)$. Therefore:
\begin{equation}
\scalebox{1.0}{$
w_s \cdot w_t = \frac{S(w_s, w_t)}{S(w_s, *) \cdot S(*, w_t)} = \frac{Dice(w_s, w_t)}{2}
$}
\label{eq:proof3}
\end{equation}
This theoretical result implies that the cross-lingual similarity function derived from embeddings based on sentence IDs is essentially a generalization of the Dice coefficient.

\subsection{SGNS with Sentence IDs}

\begin{table*}[t]
\begin{center}
\scalebox{0.8}{
\begin{tabular}{| c | c | c c || c c c | c c | }
\hline
& & & & \multicolumn{3}{c|}{\multirow{2}{*}{\bf Prior Art}} & \multicolumn{2}{c|}{\multirow{2}{*}{\bf This Work}} \\
& & & & & & & & \\
\cline{5-9}
& & & & Bilingual & Inverted & IBM & Bilingual & Multilingual \\
& & & & Autoencoders & Index & Model-1 & SID-SGNS & SID-SGNS \\
\hline
\hline
\multirow{16}{*}{\rotatebox[origin=c]{90}{\textbf{\large{Word Alignment (1-AER)}}}} & \multirow{6}{*}{\small {\sc Gra\c{c}a}} & en & fr & .4376 & .3499 & .4263 & .4167 & \textbf{.4433} \\
& & fr & en & .4488 & .3995 & .4248 & .4300 & \textbf{.4632} \\
& & en & es & \textbf{.5000} & .3443 & .4251 & .4200 & .4893 \\
& & es & en & \textbf{.5076} & .4545 & .4243 & .3610 & .5015 \\
& & en & pt & .4449 & .3263 & \textbf{.4729} & .3983 & .4047 \\
& & pt & en & .4474 & .3902 & \textbf{.4712} & .4272 & .4151 \\
\cline{2-9}
& \multirow{2}{*}{\small {\sc Hansards}} & en & fr & .4083 & .3336 & \textbf{.4360} & .3810 & .4091 \\
& & fr & en & .4218 & .3749 & \textbf{.4499} & .3806 & .4302 \\
\cline{2-9}
& \multirow{2}{*}{\small {\sc Lambert}} & en & es & .2960 & .2268 & .2400 & .2471 & \textbf{.2989} \\
& & es & en & .2905 & .2696 & .2443 & .2415 & \textbf{.3049} \\
\cline{2-9}
& \multirow{2}{*}{\small {\sc Mihalcea}} & en & ro & .2366 & .1951 & .2335 & .1986 & \textbf{.2514} \\
& & ro & en & .2545 & .2133 & .2214 & .1914 & \textbf{.2753} \\
\cline{2-9}
& \multirow{2}{*}{\small {\sc Holmqvist}} & en & sv & .2746 & .2357 & \textbf{.3405} & .2373 & .2737 \\
& & sv & en & .2994 & .2881 & \textbf{.3559} & .2853 & .3195 \\
\cline{2-9}
& \multirow{2}{*}{\small {\sc Cakmak}} & en & tr & .2256 & .1731 & \textbf{.3154} & .1547 & .2404 \\
& & tr & en & .2661 & .2665 & \textbf{.3494} & .1571 & .2945 \\
\hline
\multirow{16}{*}{\rotatebox[origin=c]{90}{\textbf{\large{Dictionary Induction (P@1)}}}} & \multirow{16}{*}{{\small \textsc{Wiktionary}}} & en & fr & .2475 & .3125 & .1791 & .3182 & \textbf{.3304} \\
& & fr & en & .2762 & .3466 & .1816 & .3379 & \textbf{.3893} \\
& & en & es & .2738 & .3135 & .0903 & .3268 & \textbf{.3509} \\
& & es & en & .3012 & .3574 & .1131 & .3483 & \textbf{.3868} \\
& & en & pt & .3281 & .3866 & .3779 & .3869 & \textbf{.4058} \\
& & pt & en & .3661 & .4190 & .4358 & .4119 & \textbf{.4376} \\
& & en & ar & .0995 & .1364 & .1316 & .1364 & \textbf{.1605} \\
& & ar & en & .1958 & .2825 & .2873 & .2408 & \textbf{.3082} \\
& & en & fi & .0887 & .1367 & .1340 & .1280 & \textbf{.1591} \\
& & fi & en & .1597 & .2477 & .2394 & .1877 & \textbf{.2584} \\
& & en & he & .0985 & .1284 & .1224 & .1403 & \textbf{.1448} \\
& & he & en & .1701 & .2179 & .2000 & .1791 & \textbf{.2403} \\
& & en & hu & .1679 & .2182 & .2219 & .2299 & \textbf{.2482} \\
& & hu & en & .2234 & .3204 & .2985 & .2759 & \textbf{.3372} \\
& & en & tr & .1770 & .2245 & .1985 & .2207 & \textbf{.2437} \\
& & tr & en & .2069 & .2835 & .3073 & .2598 & \textbf{.3080} \\
\hline
\hline
\multicolumn{4}{|c||}{\textbf{Average}} & 0.2856 & 0.2867 & 0.2922 & 0.2830 & \textbf{0.3289} \\
\multicolumn{4}{|c||}{\textbf{Top 1}} & 2 & 0 & 8 & 0 & \textbf{22} \\
\hline
\end{tabular}}
\end{center}
\caption{The performance of SID-SGNS compared to state-of-the-art cross-lingual embedding methods and traditional alignment methods.}
\label{tab:results_sgns}
\end{table*}

The Dice coefficient appears to be a particularly na\"{i}ve variant of matrix-based methods that use sentence IDs. For example, Inverted Index \cite{Soegaard2015}), which uses SVD over IDF followed by $L_2$ normalization (instead of $L_1$ normalization), shows significantly better performance. We propose using a third variant, sentence-ID SGNS (SID-SGNS), which simply applies SGNS \cite{Mikolov2013Distributed} to the word/sentence-ID matrix (see \S5.1).

Table~\ref{tab:results_sgns} compares its performance (Bilingual SID-SGNS) to the other methods, and shows that indeed, this algorithm behaves similarly to other sentence-ID-based methods. We observe similar results for other variants as well, such as SVD over positive PMI (not shown).

\subsection{Embedding Multiple Languages}
\label{sec:polyglot}

Up until now, we used bilingual data to train cross-lingual embeddings, even though our parallel corpus (the Bible) is in fact multi-lingual. Can we make better use of this fact?

An elegant property of the sentence-ID feature set is that it is a truly inter-lingual representation. This means that multiple languages can be represented together in the same matrix before factorizing it. This raises a question: does dimensionality reduction over a \emph{multi}-lingual matrix produce better cross-lingual vectors than doing so over a bilingual matrix?

We test our hypothesis by comparing the performance of embeddings trained with SID-SGNS over all 57 languages of the Bible corpus to that of the bilingual embeddings we used earlier. This consistently improves performance across all the development benchmarks, providing a 4.69\% average increase in performance (Table~\ref{tab:results_sgns}). With this advantage, SID-SGNS performs significantly better than the other methods combined.\footnote{We observed a similar increase in performance when applying the multi-lingual signal to S{\o}gaard et al.'s \shortcite{Soegaard2015} IDF-based method and to SVD over positive PMI.} This result is similar in vein to recent findings in the parsing literature \cite{Ammar2016,Guo2016}, where multi-lingual transfer was shown to improve upon bilingual transfer. 

In absolute terms, Multilingual SID-SGNS's performance is still very low. However, this experiment demonstrates that one way of making significant improvement in cross-lingual embeddings is by considering additional sources of information, such as the multi-lingual signal demonstrated here. We hypothesize that, regardless of the algorithmic approach, relying solely on sentence IDs from bilingual parallel corpora will probably not be able to improve much beyond IBM Model-1.

\section{Data Paradigms}
\label{sec:data-paradigms}

In \S\ref{sec:background}, we assumed that using sentence-aligned data is a better approach than utilizing document-aligned data. Is this the case?

To compare the data paradigms, we run the same algorithm, SID-SGNS, also on document IDs from Wikipedia.\footnote{We use the word-document matrix mined by \newcite{Soegaard2015}, which contains only a subset of our target languages: English, French, and Spanish.} We use the bilingual (not multilingual) version for both data types to control for external effects. During evaluation, we use a common vocabulary for both sentence-aligned and document-aligned embeddings.

Table~\ref{tab:data_paradigms} shows that using sentence IDs from the Bible usually outperforms Wikipedia. This remarkable result, where a small amount of parallel sentences is enough to outperform one of the largest collections of multi-lingual texts in existence, indicates that document-aligned data is an inferior paradigm for translation-related tasks such as word alignment and dictionary induction.

\begin{table}
\begin{center}
\scalebox{0.8}{
\begin{tabular}{| c | c c || c | c |}
\hline
	&		&		&	\multirow{2}{*}{\textbf{The Bible}}	&	\multirow{2}{*}{\textbf{Wikipedia}}	\\
   	&		&		&		&	\\
\hline
\hline
\multirow{4}{*}{\small {\textsc{Gra\c{c}a}}} & en & fr & \textbf{.3169} & .2602 \\
& fr & en & \textbf{.3089} & .2440 \\
& en & es & \textbf{.3225} & .2429 \\
& es & en & \textbf{.3207} & .2504 \\
\hline
\multirow{2}{*}{\small {\textsc{Hansards}}} & en & fr & \textbf{.3661} & .2365 \\
& fr & en & \textbf{.3345} & .1723 \\
\hline
\multirow{2}{*}{\small {\textsc{Lambert}}} & en & es & \textbf{.2161} & .1215 \\
& es & en & \textbf{.2123} & .1027 \\
\hline
\multirow{4}{*}{\small {\textsc{Wiktionary}}} & en & fr & .3232 & \textbf{.3889} \\
& fr & en & .3418 & \textbf{.4135} \\
& en & es & \textbf{.3307} & .3262 \\
& es & en & \textbf{.3509} & .3310 \\
\hline
\hline
\multicolumn{3}{|c||}{\textbf{Average}} & \textbf{.3121} & .2575 \\
\multicolumn{3}{|c||}{\textbf{Top 1}} & \textbf{10} & 2 \\
\hline
\end{tabular}}
\end{center}
\caption{The performance of SID-SGNS with sentence-aligned data from the Bible (31,102 verses) vs document-aligned data from Wikipedia (195,000 documents).}
\label{tab:data_paradigms}
\end{table}

\section{Conclusions}

In this paper, we draw both empirical and theoretical parallels between modern cross-lingual word embeddings based on sentence alignments and traditional word alignment algorithms. We show the importance of sentence ID features and present a new, strong baseline for cross-lingual word embeddings, inspired by the Dice aligner. Our results suggest that apart from faster algorithms and more compact representations, recent cross-lingual word embedding algorithms are still unable to outperform the traditional methods by a significant margin. However, introducing our new multi-lingual signal considerably improves performance. Therefore, we hypothesize that the information in bilingual sentence-aligned data has been thoroughly mined by existing methods, and suggest that future work explore additional sources of information in order to make substantial progress.

\section*{Acknowledgments}
The work was supported in part by The European Research Council (grant number 313695) and The Israeli Science Foundation (grant number 1555/15). We would like to thank Sarath Chandar for helping us run Bilingual Autoencoders on large datasets.

\bibliography{xling.bib}
\bibliographystyle{eacl2017}

\end{document}